\documentclass[11pt,a4paper]{article}
\usepackage[hyperref]{acl2020}
\usepackage{times}
\usepackage{latexsym}

\newcommand*\bigcdot{\mathpalette\bigcdot@{.5}}
\usepackage{soul}
\usepackage{url}
\usepackage{amssymb}
\usepackage[utf8]{inputenc}
\usepackage{graphicx}
\usepackage{amsmath}
\usepackage{booktabs}
\usepackage{algorithm}
\usepackage{algorithmic}
\usepackage{mathrsfs}
\usepackage{enumitem}
\usepackage{multirow}

\usepackage{multicol}
\usepackage{tabularx}
\usepackage{CJKutf8}
\usepackage[encapsulated]{CJK}
\usepackage{xcolor}
\usepackage{enumitem}
\usepackage{bm}
\usepackage{fixmath}
\usepackage{amsbsy}
\usepackage{tabu}

\usepackage{amsmath}
\makeatletter
\newcommand*\bigcdot@[2]{\mathbin{\vcenter{\hbox{\scalebox{#2}{$\m@th#1\bullet$}}}}}
\usepackage{microtype}

\aclfinalcopy % Uncomment this line for the final submission
\title{Response-Anticipated Memory for \\On-Demand Knowledge Integration in Response Generation}

\author{Zhiliang Tian,\thanks{\ \ This work was partially done when Zhiliang Tian was an intern at Tencent AI Lab.} \ $^{1,4}$\ \ Wei Bi,\thanks{\ \ Corresponding author} \ $^2$ \ \ Dongkyu Lee,$^{1}$ \ \  Lanqing Xue,$^{1}$ \\
\textbf{\ Yiping Song,$^{3}$ \ \ Xiaojiang Liu,$^2$ \ \ Nevin L. Zhang$^{1,4}$}\\
	\normalsize{$^1$Department of Computer Science and Engineering,}\\[-.05cm]
	\normalsize{The Hong Kong University of Science and Technology, Hong Kong SAR, China}\\[-.05cm]
	\normalsize{$^2$Tencent AI Lab, Shenzhen, China}\\[-.05cm]
	\normalsize{$^3$Department of Computer Science School of EECS, Peking University, Beijing, China} \\[-.05cm]
    \normalsize{$^4$HKUST Xiao-i Robot Joint Lab, Hong Kong SAR, China}\\[-.05cm]
	{\small\tt \{ztianac,dleear,lxueaa,lzhang\}@cse.ust.hk}  \\[-.09cm] {\small\tt \{victoriabi,kieranliu\}@tencent.com \ songyiping@pku.edu.cn}
}

\date{}

\begin{document}
\maketitle
\begin{abstract}
Neural conversation models are known to generate appropriate but non-informative responses in general. A scenario where informativeness can be significantly enhanced is Conversing by Reading (CbR), where conversations take place with respect to a given external document. In previous work, the external document is utilized by (1) creating a context-aware document memory that integrates information from the document and the conversational context, and then (2) generating responses referring to the memory. In this paper, we propose to create the document memory with some anticipated responses in mind. This is achieved using a teacher-student framework. The teacher is given the external document, the context, and the ground-truth response, and learns how to build a response-aware document memory from three sources of information. The student learns to construct a response-anticipated document memory from the first two sources, and the teacher’s insight on memory creation. Empirical results show that our model outperforms the previous state-of-the-art for the CbR task.
\end{abstract}

\section{Introduction}
Neural conversation models have achieved promising performance in response generation.
However, it is widely observed that the generated responses lack sufficient content and information~\cite{li2016diversitymmi}. %,gao2019jointly
One way to address this issue is to integrate various external information into conversation models. Examples of external information include document topics~\cite{xing2017topic}, commonsense knowledge graphs~\cite{zhou2018commonsense}, and domain-specific knowledge bases ~\cite{yang2019hybrid}.
%Recently, \newcite{CMRACL2019} study a ``conversing by reading" (CbR) scenario, in which two users converse with acquiring information from a long document.
Conversing by reading (CbR)~\cite{CMRACL2019} is a recently proposed scenario where external information can be ingested to conversations. In CbR, conversations take place with reference to a document. 
%The key challenge is to learn support on-demand knowledge integration for response generation, which is also the focus of this work.
The key problem in CbR is to learn how to integrate information from the external document into response generation on demand. 
\textbf{\begin{figure}[t]
	\centering
	\includegraphics[width=1.0\columnwidth]{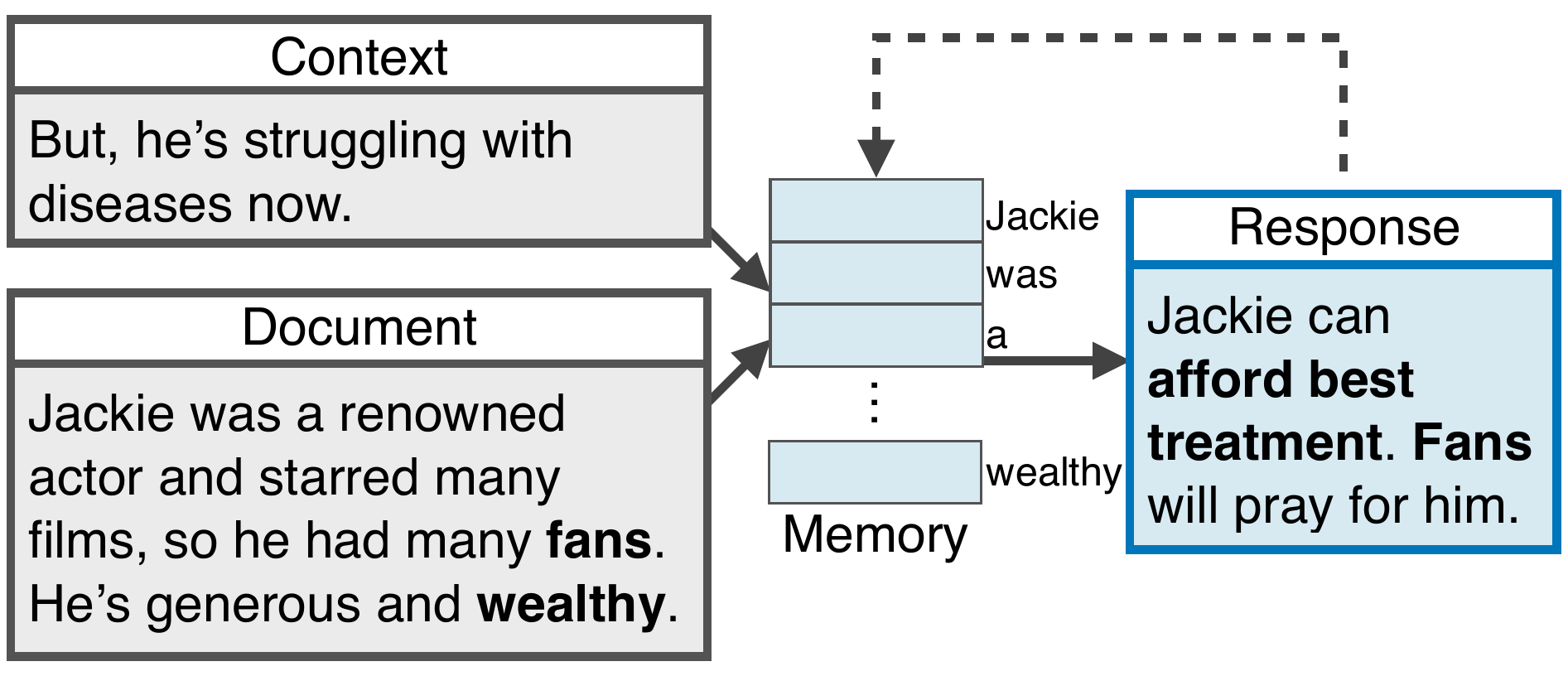}
	\caption{A motivating example of constructing a response-anticipated document memory for response generation. Details are provided in the introduction.}
\label{fig:intro_case}
\end{figure}}

To exploit knowledge from documents for conversations, a conventional way is to extend the sequence-to-sequence (Seq2Seq) model~\cite{seq2seq} with Memory Networks~\cite{sukhbaatar2015end}, which store knowledge representations accessible to their decoder~\cite{ghazvininejad2018knowledge,parthasarathi2018extending}. %\newcite{ghazvininejad2018knowledge} use an encoder-decoder model with an end-to-end memory networks \cite{e2eMemoryNet} storing knowledge facts. 
\newcite{dinan2018wizard} propose to encode the dialogue context as well as a set of retrieved knowledge by Transformer~\cite{vaswani2017attention} to construct the memory.
%``transformer memory network" to select knowledge from memory by transformer encoder, and then generate responses.
However, these methods only use sentence-level representations of the documents in the memory, which cannot pinpoint accurate token-level document information.
%Memory-based models carry abundant information, but it's hard to search from such a large memory.

To discover token-level document information,
%To strengthen the knowledge discovering from documents memory, 
researchers borrow models from other generation tasks, which are adept at extracting segments of sentences for given questions.
\newcite{moghe2018towards} explore the pointer generator network~\cite{see2017get} for abstractive summarization and the bi-directional attention flow model~\cite{seo2016bidirectional}, which is a QA model to predict a span of the document to be contained in the response. 
\newcite{CMRACL2019} follow the stochastic answer network (SAN)~\cite{SAN} in machine reading comprehension (MRC), integrating both context and document information to form the context-aware document memory. This approach obtains the state-of-the-art performance on the CbR task.
%, and generate responses assisted by that memory. This paper 
%fuse documents and contexts to select knowledge by a transformer encoder, and then to fuse knowledge and contexts to select knowledge; use another transformer to encoderd knowlege and decode to get responses.
%Previous work applies MRC systems on CbR task. \newcite{CMRACL2019} follow stochastic answer networks \cite{SAN}, the state-of-the-practice MRC model. It builds the document memory by a context-document attention and a document-document self-attention, and decodes responses from conversational context and document memory (blue part in Fig.~\ref{fig:intro_case}). Other similar tasks, like conversing with external knowledge \cite{18KGdataset_use_mrc}, also study from MRC methods \cite{seo2016bidirectional}.

However, we should notice the difference between existing generation tasks and CbR.
For summarization, QA, and MRC, they require models to extract exact answers from documents, where documents cover all requisite knowledge. 
Meanwhile, CbR expects to output a general utterance relevant to both context and document.
%, where contexts are natural utterances and documents are freeform articles written by human.
As the example in Fig.~\ref{fig:intro_case}, the document refers to \textit{actor, films, fans, wealthy} and the context mentions \textit{disease}. Document and context discuss the same person but have no topic overlap; thus we cannot pinpoint document information from the context. If we use SAN as in \newcite{CMRACL2019}, SAN can hardly acquire helpful information from context-document interaction. 
To ingest useful knowledge for response generation, we argue that processing documents should consider not only the interaction between context and document but also the target response. As in the example, the document should attend more on \textit{fans, wealthy} by considering the response.
%, so that the selected information could be more beneficial to response generation. 
%, where inputs cannot satisfy MCR's requirement.
%SAN achieves a good performance on ``conversing by reading" task, but we should notice the difference between MRC and our task.
% It can.... , but in document level, the attention among every tokens is too heavy to focus on the helpful information from the document.
%For conversation tasks, we observed some cutting-edge methods leveraging a rough response estimated by retrieval systems \cite{song2018ensemble, cai2018skeleton} or external memory \cite{MyACL2019} to augment the response generation. Similar, We believe that the estimated response may also contribute to the selection and integration in our task. 

In this work, we propose a method to construct a response-anticipated memory to contain document information that is potentially more important in generating responses. Particularly, we construct a teacher-student framework based on \citet{CMRACL2019}. The teacher model accesses the ground-truth response, context, and document. It learns to construct a weight matrix that contains information about the importance of tokens in the document to the response. 
%indicating the token importance on the document with respect to the response.
%We call the weight matrix as the response-anticipated weight matrix since it contains information about responses.
%The student model learns to mimic the response-anticipated weight matrix of the teacher, but without access to the response. 
The student model learns to mimic the weight matrix constructed by the teacher without access to the response. That is, the teacher learns to build a response-aware memory, while the student learns to build a response-anticipated memory. During inference on testing data, the student will be applied. Our experiments show our model exceeds all competing methods.
%in terms of the quality, informativeness, and the relevance to document. 

%After training, student model, which captures importance on document aware of responses, outputs responses during generation phase.
%..... (learn a graph and then conduct GAT).
%the teacher access context $x$, document $d$, and response $r$ to; the student only access $x$ and $d$.
%
% As the result, the model attends more on document tokens, which are helpful to generate responses. For the case mentioned in Fig.~\ref{fig:intro_case}, documents can attend more on \textit{fans, wealthy} due to considering responses.

%Our contributions can be summarized as followed:
%\begin{itemize}
%    \item We propose a new prospective to enhance the CbR by considering responses.
%    \item We design a novel teacher-student framework to build a response-aware document memory for CbR tasks.
%    \item Our model obtain the state-of-the-art performance on CbR tasks.
%\end{itemize}

\section{Related Work}
Most neural conversation models in open domain chit-chat scenarios are based on the Seq2Seq model~\cite{seq2seq,NRM_Shang15}.
%NRM_Shang15
%These models converse much coherently and flexibly, compared with the prior studies template-based \cite{walker2001quantitative,williams2013dialog} or retrieval-based \cite{isbell2000cobot,wang2013dataset} systems. 
%Nevertheless, such data-driven systems still highly rely on the information stored in training corpora, and suffer from generating bland, universal, and meaningless responses.
A critical issue of these models is the safe response problem, i.e., generated responses often lack enough content and information.
To address this issue, previous work encourages response diversity and informativeness by introducing new training objectives~\cite{li2016deep, zhao2017CVAE}, refining beam search strategies~\cite{li2016diversitymmi,vijayakumar2016diverse,song2017diversifying}, exploiting information from conversational contexts~\cite{aaai16_hred,serban2017hierarchical,tian2017acl}, or incorporating with retrieval-based conversation systems~\cite{song2018ensemble,wu2019response,tian2019ACL}.

%additional information including images \cite{das2017visual,mostafazadeh2017image} and knowledge bases \cite{agarwal2018knowledge,zhou2018commonsense,parthasarathi2018extending,yang2019hybri}. Others

%The involvement of external resources also enriches conversations a lot.
%Some researchers augment conversations with knowledge triplets or descriptions from knowledge bases \cite{agarwal2018knowledge,zhou2018commonsense,parthasarathi2018extending,yang2019hybrid}. Others enhance conversations by text information in different scenes.
% Some researchers augment conversations with knowledge triplets or descriptions from .
% Others enhance conversations by text information in different scenes.

Some researchers augment information in generating responses by external resources. \newcite{zhou2018commonsense} utilize the commonsense knowledge graph by their designed graph attention. \newcite{agarwal2018knowledge} propose a knowledge encoder to encode query-entity pairs from the knowledge base. \newcite{wu2019proactive} enrich response generation with knowledge triplets. %Image-grounded conversation~\cite{das2017visual,mostafazadeh2017image} is another promising direction to improve response informativeness. 
These work all uses knowledge information in structured formats.

External unstructured text information has also been investigated to improve conversation models.
Some researchers directly build ``document memory" by using distributed representations of the knowledge sentences into conversation models~ \cite{ghazvininejad2018knowledge,parthasarathi2018extending}.
\newcite{dinan2018wizard} make use of the Transformer~\cite{vaswani2017attention} to encode the knowledge sentences as well as the dialogue context. 
\newcite{ren2019thinking} design a knowledge selector to construct the document memory on selective knowledge information.
%\newcite{gopalakrishnan2019topical}
As stated in the introduction, some other researchers borrow models from other generation tasks, including abstractive summarization models~\cite{moghe2018towards}, QA models~\cite{moghe2018towards} and MRC models~\cite{meng2019refnet,CMRACL2019}.
%
%For MRC-based models, \newcite{moghe2018towards} study from an MRC model \cite{seo2016bidirectional}, while \newcite{meng2019refnet} amend the copy mechanism in MRC. 
Especially, \newcite{CMRACL2019} get the state-of-the-art performance.
%All the above models can hardly handle the case mentioned in Fig.~\ref{fig:intro_case}, since
However, they all construct the document memory relying on connections between context and document without consideration of the response.  If context or document contains a lot of noise tokens irrelevant to the response, which is indeed the case in CbR, the constructed memory may be misled by these noise information (as the case in Fig.~\ref{fig:intro_case}).
Therefore, we propose to involve the consideration of responses in the memory construction, which can benefit generating a more desired response.
%Based on their models, our work attends more on potential responses, which is more helpful in conversation scenarios.
%We choose three representative models \cite{ghazvininejad2018knowledge}, \cite{ren2019thinking}, and \cite{CMRACL2019} as our baselines.
%Compared to CVAE,

%utilize domain specific data like movie plot and fact table on conversations related to movie domain. 
%\newcite{gopalakrishnan2019topical} construct a dataset that provides a structured ``reading set" for closed domain conversations. 
%propose a new task that teaching knowledge to students by conversations, with the purpose of making conversations to cover the knowledge. Different from above scenarios, our task focuses on open domain chit-chat and exploits information only from raw document directly. 

%\newcite{CMRACL2019} propose conversing by reading (CbR) task and follow a  MRC framework \cite{SAN} to construct their model, which is the closest work to ours. Our paper propose to consider responses information when reading documents based on their work.

\section{Methodology}
In this section, we will first give an overall description of the proposed teacher-student architecture for CbR, then briefly describe the base model. The detailed teacher model and student model are presented in Sec~\ref{sec:teacher_model} and~\ref{sec:student_model}. Lastly, we summarize the training updates of the two models in Sec~\ref{sec:model_training}.

\subsection{Model Architecture}
%\footnote{***should give a brief motivation.}  
%\footnote{***u have not yet introduced what is doc mem here. also, what does the selector G here means?}
%\footnote{***should give an intuitive explanation.}
The CbR task provides a conversation context %$X$=$\{x_1,x_2,...,x_{|X|}\}$ 
$X$ and a document $D$
as inputs, requiring the model to generate a response $R$
to $X$ by referring to $D$. In the rest of the paper, we use $| X|$, $|D|$, and $|R|$ to denote the number of tokens in $X$, $D$, and $R$ respectively.
%=$\{d_1,d_2,...,d_{|D|}\}$ 
%=$\{r_1,r_2,...,r_{|R|}\}$ 
%\footnote{***this sentence is not motivated enough.}
% To pick proper information from document respecting to responses, 
To pinpoint accurate document information for response generation, we design a teacher-student framework to construct document memory as follows: 
%to let teacher detect and utilize the correlation between document and response, and force student to approach its teacher.
%\footnote{***will leave this part as it is now at this point. need to revise it after editing the whole model section.}
\begin{itemize}[wide=0\parindent,noitemsep]
    %for response generation, with the supervision signals constructed as a ``selecting graph'' $\mathcal{G}$ which has 
    %with access to the response.
    %while discovering the information, which is helpful to response generation, from the document. Its inputs are the document, the conversational contexts, and the response; the output is the response. To discover the helpful information, we propose to permit the teacher model to
    %
    %is packed with an attentional recurrent decoder.
    %A standard memory matrix is constructed to capture the ingested information from the dialogue context and the document. 
    %For each decoding step in our decoder, we generate a response token by attending to the memory $\mathbf{M}_{orig}$.
    %That is, we construct the memory
    %access the ground-truth response. It 
    %by glancing at the response as well as the dialogue context and the document. 
    \item The teacher model learns a response-aware document memory $\mathbf{M}$ used in our base conversation model. Specifically, we construct a response-aware weight matrix $\mathbf{G} \in \mathbb{R}^{|D| \times |D|}$, which considers the correlation between context-aware document representations and response representations, and then impose $\mathbf{G}$ on the memory matrix $\mathbf{M}$.
    The teacher model is optimized to reconstruct the response with the use of response-aware memory $\mathbf{M}$.
    %(yellow parts in Fig~\ref{fig:intro_case} \& Fig~\ref{}). 
    \item The student model learns to construct a response-anticipated weight matrix to estimate $\mathbf{G}$ used in the teacher model but without access to the response. It is a feed-forward neural network with document and context as its input. 
    %\footnote{***add a sentence to describe the model of student model.}
    %and then generates responses. We construct a ``study network" to learn $\mathcal{G}$ from teacher.
\end{itemize}
The teacher model and the student model are jointly optimized with training data, while only the student model is applied to testing data.

%\begin{figure}[htb]
%	\centering
%	\includegraphics[width=0.45\textwidth]{modelskeleton}
%	\caption{The skeleton of our model.}
%	\label{fig:modelskeleton}
%\end{figure}

\subsection{Base Model}
\label{sec:base_model}
Following~\citet{CMRACL2019}, we use SAN~\cite{SAN} as our base model, which mainly consists of three components: 
%to encode both the context $X$ and the document $D$, fuse their information into a memory matrix, and then forward it to a response generator.
%response $r$ given $d$ and $x$ 
%(blue parts in Fig.~\ref{fig:intro_case} \& Fig.~\ref{}). 
% \begin{flalign}
% \label{eq:base_enc}
% & \mathbf{D}^0 = \mbox{Encoder}_d(d) && \mathbf{X} = \mbox{Encoder}_x(x) \\
% \label{eq:base_attn}
% & \mathbf{D} = \mbox{CrossAttn}(\mathbf{D}^0, \mathbf{X}) && \mathbf{M} = \mbox{SelfAttn}(\mathbf{D}) \\
% \label{eq:base_dec}
% & \hat{r} = \mbox{Decoder}(\mathbf{X}, \mathbf{M}) && 
% \end{flalign}

\begin{itemize}[wide=0\parindent,noitemsep]
    \item Input encoder: We use two bi-directional LSTM encoders to extract token-level representations of the document $D$ and the context $X$.
    %, denoted as
%$\mathbf{D}=[\mathbf{d}_1,\ldots, \mathbf{d}_{|D|}]$ and $\mathbf{X}=[\mathbf{x}_1,\ldots, \mathbf{x}_{|D|}]$ 
%(Eq.~\ref{eq:base_enc}), 
%where $|D|$ and $|X|$ are the length of the document and context respectively. 
%$$and dimension of representations respectively.
%Each encoder involves a word embedding layer, a fully-connected layer, and a 
%We use bi-directional LSTM here. 
    \item Memory construction: We build the document memory $\mathbf{M} \in \mathbb{R}^{|D| \times k}$ ($k$ is the hidden size of the memory) which will be used in the decoder.
 %to store all token-level representations in the current document, where
  A cross-attention layer is first applied to the outputs of the two encoders to integrate information from the context to the document. Then, we obtain a set of context-aware document representation $\mathbf{D}=[\mathbf{d}_1,\ldots, \mathbf{d}_{|D|}]$. Since each $\mathbf{d}_i$ corresponds to a document token, we treat it as the context-aware token representation of the $i$-th token. 
 %to obtain contextual the token-level representation of context and document, and 
 Next, a self-attention layer is employed to ingest salient information of the context-aware document representations:
 %\begin{equation}
 %    \mathbf{M} = \mbox{SelfAttn}(\mathbf{D}) = \mbox{softmax} (\mathbf{D}^T \mathbf{D})  \mathbf{D}^T.
 %\end{equation} 
  \begin{equation}
  \label{eq:base_attn}
     \mathbf{M} = \mbox{SelfAttn}(\mathbf{D}) = \mathbf{A} \mathbf{D}^T,  \mathbf{A}=\mbox{softmax} (\mathbf{D}^T \mathbf{D}) 
 \end{equation} 
 where the softmax conducts the normalization over each row of the matrix.
 %(Eq.~\ref{eq:base_attn}).
    \item Output decoder: We use an attentional recurrent decoder to generate response tokens by attending to the memory $\mathbf{M}$. The initial hidden state is set as the summation of token-level context representations. For each decoding step $t$, we get a hidden state $\mathbf{h}_t$:
    \begin{eqnarray}
    \label{eq:base_dec1}
    %\begin{aligned}
     \mathbf{z}_t & = & \mbox{GRU}(\mathbf{e}_{t-1}, \mathbf{h}_{t-1}), \\
     \label{eq:base_dec2}
     \mathbf{h}_t & = & \mathbf{W_1} [\mathbf{z}_t; \mbox{CrossAttn}(\mathbf{z}_t,\mathbf{M})]
     %p(y_t) & = &  \mbox{softmax}(\mathbf{W_2}\mathbf{h}_t + \mathbf{b}), 
    %\end{aligned}
    \end{eqnarray}
    where $[;]$ indicates concatenation, and the cross-attention layer here integrates information from the memory to the recurrent outputs.
    %\footnote{***do not understand the output of this operation}
    %; 
    $\mathbf{e}_{t-1}$ is the word-embedding at step $t-1$. Finally, we generate a token $y_t$ by a softmax on $\mathbf{h}_t$.
    % The word embeddings, $\mathbf{W_1}$, $\mathbf{W_2}$, and $\mathbf{b}$ are decoder parameters.
    
\end{itemize} 
Our model modifies the memory construction by refining its self-attention layer so that the memory represents more accurate and on-demand knowledge that helps generating the response.
%cells which are  helpful to response generation. % the correlation of documents and responses,
%First, 

%Then, 
%The decoder treats the summation of context tokens $\sum_{i=0}^{|x|} \mathbf{X}_i$ as its initial state, and then it exploits the document memory by a cross attention between document and each hidden state in decoding (Eq.~\ref{eq:base_dec}).

% \begin{figure*}[htb]
% 	\centering
% 	\includegraphics[width=1\textwidth]{model.png}
% 	\caption{The architecture of our model with a case. Gray lines and blocks indicate the  base model; the combination of gray and blue parts indicates teacher model; the combination of gray and yellow parts means student model. In (estimated) selecting graph and memory, shaded area shows the high weight and no shaded means low weight. For example, \textit{Jackie}, \textit{renowned} and \textit{wealthy} have high weights on selecting graph, resulting in high weight on memory.}
% 	\label{fig:model}
% \end{figure*}

\begin{figure*}[htb]
	\centering
	\includegraphics[width=1\textwidth]{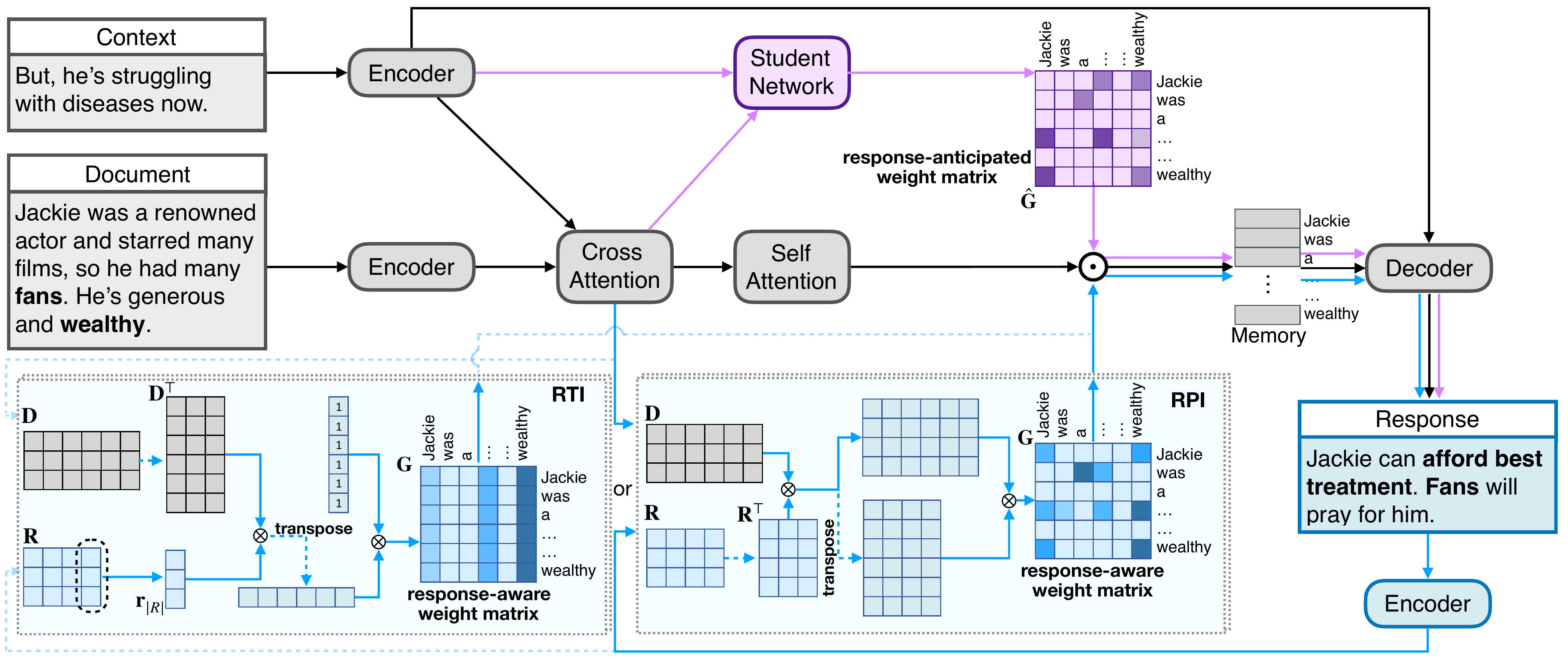}
	\caption{The architecture of our model. Blocks and lines in gray color compose the base model. Blue and gray parts compose the teacher model, while purple parts compose the student model. All components work for training, while only the student model and the decoder works for inference. In the response-aware/anticipated weight matrix, darker grids indicate higher weights. $(\bigotimes$: matrix multiplication; $\bigodot$: element-wise matrix multiplication.$)$
	%For example, \textit{Jackie}, \textit{renowned} and \textit{wealthy} have high weights in the weight matrix.
	%During training, the teacher and student models will be optimized jointly. In testing, only the student model will be used to infer the response-anticipated weight matrix and the decoder will utilize it to generate the output response.
	}
	\label{fig:model}
\end{figure*}

\subsection{Teacher Model}
\label{sec:teacher_model}
To ingest accurate memory information for response generation under the aforementioned base model, our teacher model builds a response-aware weight matrix $\mathbf{G} \in \mathbb{R}^{|D| \times |D|}$ given the context-aware document representation $\mathbf{D}$ and the response $R$, then refines the document memory $\mathbf{M}$ with $\mathbf{G}$.
%(yellow parts in Fig~\ref{fig:intro_case} \& Fig~\ref{}). 
%\footnote{***first sentence needs to restate the motivation of why looking at the response.}
%
Elements in $\mathbf{G}$'s indicate the importance of tokens or token pairs in the document, with consideration of the response information. 

First, we describe how to modify the memory matrix $\mathbf{M}$ when $\mathbf{G}$ is given. The original memory $\mathbf{M}$ is constructed by a self-attention operation as Eq.~\ref{eq:base_attn}. To facilitate response awareness, we update the attention weight matrix $\mathbf{A}$ by element-wise multiplying $\mathbf{G}$, and then get the refined memory $\mathbf{\widetilde{M}}$ as
\begin{equation}
%\begin{flalign}
\label{eq:overall_refine}
 \mathbf{A} = \mbox{softmax} (\mathbf{D}^T \mathbf{D}),  %\mathbf{\widetilde{A}} = \mathbf{G} \odot \mathbf{A}, 
  \mathbf{\widetilde{M}} = (\mathbf{G} \odot \mathbf{A}) \mathbf{D}^T.
%\end{flalign}
\end{equation}
% As can be seen, in a standard self-attention layer, $\mathbf{G}$ can be set as a matrix with all elements as 1.
%
In the following, we describe two methods to construct the response-aware weight matrix $\mathbf{G}$:
(1) 
%Construction by response-anticipated token importance (RTI):
We measure the response-aware token importance (RTI) considering the ground-truth response to construct $\mathbf{G}$. 
% We measure the relevance between document token $i$ and the response and treat the relevance as the importance weight of token $i$. The importance of token $i$ serves as the weight of row $i$ in $\mathbf{G}$.
(2)	
%Construction by response-anticipated token-pair importance (RPI): 
We measure the response-aware pairwise importance (RPI) of each token pair $(i, j)$, which can be directly assigned to the element $G_{ij}$ in $\mathbf{G}$. 
%We take response into consideration when measuring the importance of token $i$ and $j$ (weight of element $G_{ij}$).
For both methods, matrix elements can be either continuous or binary.

%, we describe two methods to construct the response-anticipated weight matrix $\mathbf{G}$:(1) construction by response-relevance of tokens. We calculate relevance of token $i$ to the ground-truth response, and assign the relevance as the 
%we assign an importance weight of each node in $\mathcal{V}$ to construct $\mathbf{G}$; (2)	construction by response-mediated token relations.
% in edge weighted graph, we calculate the importance weight of each edge in $\mathcal{E}$ to derive $\mathbf{G}$. 
%\footnote{***give a brief desc of each method below.}

%we propose two morphs of graphs as following:

\noindent\textbf{Response-Aware Token Importance (RTI)}
\\
%\noindent
We denote the response-aware token importance of document tokens as $\boldsymbol{\beta} \in \mathbb{R}^{|D|}$, and measure it by response $R$ and context-aware token representation $\mathbf{D}$. %Here, we calculate $\boldsymbol{\beta}$ to construct $\mathbf{G}$
%and use token importance to construct $\mathbf{G}$.
%In this method, we will assign a weight on each context-aware document representation.
%\footnote{***change the first sentence}
%row in $\mathbf{G}$ and denote $\boldsymbol{\beta} \in \mathbb{R}^{|D|}$ as the weighting vector.
To obtain $\boldsymbol{\beta}$, we first apply an encoder to obtain the token-level representations of the response as $[\mathbf{r}_{1},\ldots,\mathbf{r}_{|R|}]$ and use its last hidden state $\mathbf{r}_{|R|}$ as the sentence-level response representation.
%In this method, we will assign a weight on row in $\mathbf{G}$ and denote $\boldsymbol{\beta} \in \mathbb{R}^{|D|}$ as the weighting vector.We first apply an encoder to obtain the token-level representations of the response as $[\mathbf{r}_{1},\ldots,\mathbf{r}_{|R|}]$ and
%We use its last hidden state $\mathbf{r}_{|R|}$ as the response sentence-level representation:
%\begin{equation}
%\centering
%\label{eq:response_enc}
%[\mathbf{r}_1,\ldots, \mathbf{r}_{|R|}] = %\mbox{Encoder}_{R}(R).
%\end{equation}
The response-aware token importance of token $i$ is defined as the similarity between its context-aware token representation $\mathbf{d}_i$ and the response representation $\mathbf{r}_{|R|}$.
%Each node will be assigned with a weight $\beta_{i}$, which is defined as the correlation between each context-aware token representation $\mathbf{d}_i$ and the response representation $\mathbf{r}_{|R|}$. 
Next, we adjust each attention distribution (i.e., each column of $\mathbf{A}$) with each of its attention weight multiplied by the token importance $\beta_i$. 
Therefore, the resulting $\mathbf{G}$ can be obtained as:

\begin{equation}
\centering
\label{eq:soft_node_weight}
%&& \mathbf{R} = [\mathbf{r}_1,\ldots, 
%\mathbf{r}_{|R|}] = \mbox{Encoder}(R), \\
\beta_i = \mathbf{d}_i^T \mathbf{r}_{|R|}, \quad \mathbf{G} = \mathbf{1} \boldsymbol{\beta}^T,  
%&&\mathbf{A} = \mbox{softmax} (\mathbf{D} \mathbf{D}^T) \\
%&&\mathbf{\bar{A}}_{i,j} = \beta_i \cdot \mathbf{A}_{ij} \quad \forall i,j\\
%&&\mathbf{\bar{M}} = \mathbf{\bar{A}} \mathbf{D}
\end{equation}
where $\mathbf{1} \in \mathbb{R}^{|D|}$ represents an identity vector with all elements as $1$. 
%For each node to be emphasized, it increases the attention weights of all out-edges connected to that token.
By plugging the above $\mathbf{G}$ in Eq.~\ref{eq:soft_node_weight}, we can construct a memory matrix with plagiarized signals from the response.
In this way, 
the self-attention distributions can adjust to emphasize important tokens, and their corresponding context-aware document token representations become more important in the memory matrix.
%After obtaining the selecting graph $\mathcal{G}^n$, we upgrade the document memory $\mathbf{M}$ by highlighting the information on important nodes and weakening unimportant ones. 
%The base model constructs $\mathbf{M}$ by self-attention over all document tokens (Eq.~\ref{eq:self_attn}), where the self-attention can be regarded as a completed graph $\mathcal{K}$ (tokens as nodes; edges as attention weights). Notice that, graph $\mathcal{K}$ and $\mathcal{G}^n$ are isomorphic graphs, so we can update $\mathcal{K}$'s attention weights by $\mathcal{G}^n$.

%We apply two updating strategies: \textit{Soft Selection} refines self-attention by multiplying the node weight. It updates the attention weight $\boldsymbol{\alpha}$ to $\boldsymbol{\alpha}^{n}$ by element-wise multiplying node weight $\boldsymbol{\beta}$, and then gets memory by summing up $\mathbf{D}$ with the weight $\boldsymbol{\alpha}^{n}$ (Eq.~\ref{eq:node_filter}). 

Recall that the document contains a large amount of noise information in CbR. Thus the attention distributions may become long-tailed due to the existence of many redundant document tokens.
%\textit{Hard Selection} refines self-attention by 
Hence, we can further construct a binary weighting vector based on $\boldsymbol{\beta}$.
%filtering out unimportant nodes according to the node weights. 
We keep the weight of each element as 1 with the probability of $\beta_i$ calculated in Eq.~\ref{eq:soft_node_weight}.
If the weight of a token turns to 0, this token is deactivated in calculating the attention distributions. 
%It transforms node weights to binary variable to indicate whether using the node. 
However, the binary weight sampled from the Bernoulli distribution is not differentiable. To enable back-propagation of our model, we apply the Gumbel-Softmax \cite{jang2016categorical} to approximate the Bernoulli distribution in the training phase, and sample the binary value from the Bernoulli distribution in the prediction phase as: 
\begin{equation}
%\begin{aligned}
 \mathbf{G} = \mathbf{1} g(\boldsymbol{\beta})^T, \label{eq:hard_node_weight}
 \end{equation}
 where $g(\boldsymbol{\beta})$ is defined as:
\begin{eqnarray}
\left\{\begin{array}{lr}
 g(\beta_i)  = \mbox{GumbelSoftmax}(\beta_i)& \mbox{Training}, \\
 g(\beta_i) \sim \mbox{Bernoulli}(\beta_i) & \mbox{Prediction}.
\end{array} \right.\!\!\!
%\end{aligned}
%\right.
\end{eqnarray}
%\begin{flalign}
%\label{eq:gumbel2}
%& g(\beta_i)=\left\{
%\begin{aligned}
%&  \mbox{GumbelSoftmax}(\beta_i) & &\mbox{Training}, \\
%&  \mbox{Bernoulli}(\beta_i) & & \mbox{Prediction}.     
%\end{aligned}
%\right.
%\end{flalign}
%where $g(\cdot)$ outputs a binary variable or a probability for $\boldsymbol{\beta}$'s each element $\beta_i$.
\noindent
The objective function of the teacher model is to maximize the log-likelihood of responses generated by the response-aware memory constructed with $\boldsymbol{\beta}$:
\begin{equation}
%\begin{flalign}
\label{eq:teacher_loss}
 \boldsymbol{\beta} = f^t_{\theta_t}(D,X,R), 
 \mathcal{J}_t = \!\!\!\!\!\mathop{\mathbb{E}}_{_{D,X,R \sim \mathcal{D}}}\!\!\! \log P_{\phi}(R|D,X,\boldsymbol{\beta}),
%\end{flalign}
\end{equation}
where $f^t$ denotes operations in Eq.~\ref{eq:soft_node_weight} and its preorder operations. $\theta_t$ consists of all parameters in the layers of $f^t$. $\phi$ denotes parameters in Eq.~\ref{eq:base_attn} to Eq.~\ref{eq:base_dec2}. Both $\phi$ and $\theta_t$ are learning parameters for $\mathcal{J}_t$.

\noindent\textbf{Response-Aware Pairwise Importance (RPI)}
\\
%\noindent
Instead of using token importance, we can construct $\mathbf{G}$ by the pairwise importance of token pairs.
%. Thus, the edge between node $i$ and node $j$ in $\mathcal{G}$ can be associated with a weight $\mathbf{B}_{ij}$ defined as the inner-product between $\mathbf{n}_i$ and $\mathbf{n}_j$.
%can also carry weights on its edges, denoted by $\mathcal{G}^e$. 
After obtaining the token representations $[\mathbf{r}_1, \dots,\mathbf{r}_{|R|}]$ from the response encoder similarly as in RTI,
we can calculate the similarity of each $\mathbf{d}_i$ towards all $\mathbf{r}_j$'s, denoted as $\mathbf{n}_i \in \mathbb{R}^{|R|}$.
%If two nodes in $\mathcal{G}$ share similar  $\mathbf{n}_i$'s, they should be considered highly-correlated.
%Thus, the edge between node $i$ and node $j$ in $\mathcal{G}$ can be associated with a weight $\mathbf{B}_{ij}$ defined  %(element of $\mathbf{G}$)  as the inner-product between $\mathbf{n}_i$ and $\mathbf{n}_j$.
Each element in $\mathbf{G}$ can be associated with a weight $\mathbf{B}_{ij}$ defined as the inner-product between $\mathbf{n}_i$ and $\mathbf{n}_j$. Thus, we can treat $\mathbf{B}$ as the response-aware pairwise importance, and directly set each element in $\mathbf{G}$ as $\mathbf{B}_{ij}$:%(Eq.~\ref{eq:edge_weight}).
%
%Similar to the processing on $\mathcal{G}^n$, \textit{Soft Selection} updates the attention weight $\mathbf{A}$ to $\mathbf{\widetilde{A}}$ by element-wise multiplying $\mathbf{G}$, and sums up $\mathbf{d}_j$ weighted by $\boldsymbol{\alpha}^{e}$ to obtain the memory (Eq.~\ref{eq:edge_filter}). \textit{Hard Selection}
%
\begin{equation}
\label{eq:edge_weight}
\mathbf{n}_i = [\mathbf{r}_1,\ldots, \mathbf{r}_{|R|}]^T \mathbf{d}_i, \mathbf{B}_{ij} =  \mathbf{n}_i^T  \mathbf{n}_j, \mathbf{G} = \mathbf{B}.
\end{equation}
%
%In summary, our teacher model leverages the correlation between documents and responses to refine the document memory $\mathbf{M}$. To achieve it, the model updates documents' self-attention by the selector $\mathcal{G}^n$ or $\mathcal{G}^e$. 
%
% $\mathcal{G}^e$ is aware of the importance on both tokens and the correlation of token pairs.
Compared with response-aware token importance in which the designed $\mathbf{G}$ has identical column values, response-aware pairwise importance allows different values of different index $(i,j)$'s in $\mathbf{G}$ (but $(i,j)$ and $(j,i)$ have the same value since $\mathbf{G}$ is symmetric). Thus, the space of $\mathbf{G}$ is larger. % Each edge captures the importance of correlation between tokens, while the summarization of all edges adjacent to a token indicates the importance of token.

Notice that, the aforementioned binary processing with each $\beta_i$ can also be applied on each $\mathbf{B}_{ij}$ here and the resulting $\mathbf{G}$ is binary.
By using a binary $\mathbf{G}$ in our model, the memory construction can be considered as passing through a Graph Attention Network (GAT)~\cite{GAT2017}, which also constructs a graph and updates its representations relying on the information from itself and neighbors on the graph. However, our neighborhood matrix (i.e. $\mathbf{G}$ in our model) is not pre-defined as in GAT but dependant on the inputs $\mathbf{d}_i$'s and $\mathbf{r}_j$'s, which involve parameters to be estimated. 

The objective of the teacher model for RPI can be modified from Eq.~\ref{eq:teacher_loss} by replacing $\boldsymbol{\beta}$ with $\mathbf{B}$ obtained in Eq.~\ref{eq:edge_weight}.

% access the ground-truth response $r$ 
%to upgrade the document memory. Plagiarized attention
%, so that teacher would focus more on signals in documents, which is helpful for response generation.

% reveals which signals are helpful to response generation due to its accessibility of $r$.

\subsection{Student Model}
\label{sec:student_model}
The student model learns to construct a response-anticipated weight matrix to estimate the weight matrix $\mathbf{G}$ in the teacher model without access to the ground-truth $R$.
%As the lower branch of Figure~\ref{}, we construct a ``study network" to estimate $\mathbf{G}$ given the representations of documents $\mathbf{D}$ and context $\mathbf{X}$ only.
If we employ RTI, the estimated target of the student model is $\boldsymbol{\beta}$ in Eq.~\ref{eq:soft_node_weight}. For RPI, the estimated target is $\mathbf{B}$ in Eq.~\ref{eq:edge_weight}.

Given $\mathbf{D}$ and $\mathbf{X}$ as inputs,
we apply a bilinear attention layer to obtain a hidden representation matrix $\mathbf{H}$.  %(Eq.~\ref{eq:bilinear_attn})
%which can handle the input and output in variable-length and contains some learnable parameters. 
We apply a two-layer multi-layer perceptron (MLP) with ReLU activation to estimate $\boldsymbol{\beta}$; 
%then extends $\hat{\boldsymbol{\beta}}$ to $\hat{\mathbf{G}}$ in the same way of the teacher (Eq.~\ref{eq:soft_node_weight} \& ~\ref{eq:hard_node_weight}). 
% where the MLP consist of two parameters $\mathbf{W_a} \in \mathbb{R}^{k \times k}$
%, $\mathbf{W_b} \in \mathbb{R}^{k \times 1}$, and a $ReLU$ activation function.
we combine two attention outputs by $\mathbf{W_a}$ to estimate $\mathbf{B}$ in the RPI:
\begin{eqnarray}
& \mathbf{H}  =  \mbox{softmax}(\mathbf{D}^T \mathbf{W} \mathbf{X})  \mathbf{X}^T,  \label{eq:bilinear_attn}\\
&\!\!\!\left\{
\begin{array}{ll}
 \hat{\boldsymbol{\beta}}  =  \mbox{MLP}(\mathbf{H}) & \mbox{for RTI},\\
\label{eq:edge_student}
 \hat{\mathbf{B}}  =  \mathbf{H} \mathbf{W_a}\mathbf{H}^T &  \mbox{for RPI.}
 \end{array}\right.
\end{eqnarray}
The objective function of the student model is to maximize the log-likelihood of generating responses based on the estimated $\hat{\boldsymbol{\beta}}$ or $\hat{\mathbf{B}}$, and diminish the gap of the weighting vector or matrix between the student model and the teacher model by a mean square loss.
Taking the RTI strategy as an example, we optimize the following objective:
\begin{eqnarray}
%\begin{flalign}
\label{eq:student_loss}
 &\hat{\boldsymbol{\beta}} & =   f^s_{\theta_s}(D,X), \\
 & \mathcal{J}_s &  =  \!\!\!\!\!\!\!\!\mathop{{}\mathbb{E}}_{_{D,X,R\sim \mathcal{D}}} \!\!\!\!\!\!\!\log P_{\phi}(R|D,X,\hat{\boldsymbol{\beta}}) \! - \!\lambda \mathop{{}\mathcal{L}_{_{\text{MSE}}}}(\boldsymbol{\beta}, \hat{\boldsymbol{\beta}}),\nonumber
%\end{flalign}
\end{eqnarray}
\noindent
where $f^s$ denotes the operation in Eq.~\ref{eq:edge_student} and its preorder operations. $\theta_s$ consists of the layer parameters in $f^s$.
$\lambda$ balances the two loss terms. 
For RPI, we replace to optimize with $\mathbf{B}$ and $\hat{\mathbf{B}}$.
%gets the function for edge weight graph.

\begin{table*}[htb]
\centering
\scriptsize
\begin{tabular}{c|c|c|c|c|c|c|c|c|c|c|c|c|c}
\hline
\multirow{2}{*}{} & \multicolumn{3}{c|}{Appropriateness}               & \multicolumn{6}{c|}{Grounding}                      & \multicolumn{3}{c|}{Informativeness}             &      \\ \cline{2-13}
                  & NIST           & BLEU            & METEOR          & P       & R          & F1 & P$_{GT}$       & R$_{GT}$          & F1$_{GT}$             & Ent-4           & Dist-1          & Dist-2          & Len  \\ \hline \hline
Human             & 2.650          & 3.13\%          & 8.31\%          & 2.89\%          & 0.45\%          & 0.78\%  & 0.44\%          & 0.09\%          & 0.14\%          & 10.445         & 0.167          & 0.670          & 18.8 \\ \hline
Seq2Seq           & 2.223          & 1.09\%          & 7.34\%          & 1.20\%          & 0.05\%          & 0.10\%  & 0.89\%          & 0.05\%          & 0.09\%        & 9.745          & 0.023          & 0.174          & 15.9 \\ \hline
MemNet           & 2.185          & 1.10\%          & 7.31\%          & 1.25\%          & 0.06\%          & 0.12\%  & 0.91\%          & 0.05\%          & 0.10\%        & 9.821          & 0.035          & 0.226          & 15.5 \\ \hline
GLKS              & 2.413          & 1.34\%          & 7.61\%          & 2.47\%          & 0.13\%          & 0.24\% & 0.84\%          & 0.05\%          & 0.10\%         & 9.715          & 0.034          & 0.213          & 15.3 \\ \hline
CMR              & 2.238          & 1.38\%          & 7.46\%          & 3.39\%          & 0.20\%          & 0.38\%   & 0.91\%          & 0.05\%          & 0.10\%       & 9.887          & 0.052          & 0.283          & 15.2 \\ \hline
CMR+Copy          & 2.155          & 1.41\%          & 7.39\%          & 5.37\%          & 0.28\%          & 0.54\%   & 0.92\%          & 0.06\%          & 0.11\%       & 9.798          & 0.044          & 0.266          & 14.4 \\ \hline \hline
RAM\_T       & \textbf{2.510} & \textbf{1.43\%} & \textbf{7.74\%} & 4.46\%          & 0.26\%          & 0.49\%    & \textbf{1.04\%}          & \textbf{0.08\%}          & \textbf{0.15\%}      & \textbf{9.900} & \textbf{0.053} & \textbf{0.290} & 15.1 \\ \hline
RAM\_P       & 2.353          & 1.40\%          & 7.59\%          & 3.89\%          & 0.21\%          & 0.41\%    & 0.97\%          & 0.07\%          & 0.13\%      & 9.891          & 0.049          & 0.279          & 14.9 \\ \hline
RAM\_T+Copy  & 2.467          & 1.41\%          & 7.64\%          & \textbf{6.14\%} & \textbf{0.32\%} & \textbf{0.61\%} & 0.65\%          & 0.04\%          & 0.08\% & 9.813          & 0.045          & 0.265          & 14.9 \\ \hline
RAM\_P+Copy  & 2.342          & 1.41\%          & 7.51\%          & 5.83\%          & 0.30\%          & 0.57\%      & 0.84\%          & 0.06\%          & 0.10\%    & 9.798          & 0.045          & 0.267          & 14.6 \\ \hline
\end{tabular}
\caption{Automatic evaluation results on all competing methods. \textit{Len} denotes the length of the generated responses.}
\label{tb:overall_auto}
\end{table*}

\subsection{Model Training}
\label{sec:model_training}
%\footnote{***better put the loss of the teacher/student model in the last of each of their sec.}
%As the upper branch in Figure~\ref{fig:modelskeleton} and loss function in Eq.~\ref{eq:teacher_loss}, teacher model obtains selector $\mathcal{G}$ given $d$ and $d$, and then generates the response $\hat{r}$ with $d$, $x$, and $\mathcal{G}$. As the lower branch in Figure~\ref{fig:modelskeleton} and loss function in Eq.~\ref{eq:student_loss}, the student model estimate $\hat{\mathcal{G}}$ given $d$ and $x$, and generates the $\hat{r}$ from $d$, $x$, and $\mathcal{G}$. Teaching loss (Eq.~\ref{eq:teaching_loss}) helps students to diminish  $\mathcal{G}$-$\hat{\mathcal{G}}$ gap.
%

% We train the teacher model, and then switch to train the student model  .
% To enable the memory module and the generative model to work together, we combine and jointly train them. 
%\footnote{***poor writing. dont understand.}
We first train the teacher model until it converges, and then train the student model with the use of $\boldsymbol{\beta}$ or $\mathbf{B}$ from the converged teacher model. Next, we repeat the above processes iteratively.
%by a few epochs.
%
%We separate the training of the teacher model and the student model into two phases, and 
%We train the two models iteratively. 
%For the first phase, we train the teacher model. 
%The two phases switch once the model converges in current phase. 
In the training of the teacher model, we fix parameters in $\theta_s$ (except parameters shared with $\theta_t$) and train the model subject to $\mathcal{J}_t$; for the student model, we fix $\phi$ and $\theta_t$ (except parameters shared with $\theta_s$) and train the model subject to $\mathcal{J}_s$. For inference, only the student model will be used to infer the response-anticipated weight matrix and the decoder applies it for generating the output response.

%\footnote{***discuss the flexibility of the two methods vs. their training difficulty of each method here.}
As stated in RPI, it has better model capacity by allowing a larger space of $\mathbf{G}$ with the use of the weight matrix $\mathbf{B}$ instead of the token importance vector $\boldsymbol{\beta}$ in RTI.
In terms of optimization, we need to estimate more parameters by using RPI, which requires higher training difficulty.
%Compared with node weighted graph, edge weighted graph has a higher capacity ($|D| \times |D|$ free variables) to carries more information. In edge weighted graph, all edge weights connected to a token also indicate importance of that token. Hence, edge weighted graph expresses not only the token importance but also correlation between tokens. Nevertheless, it is hard for the student model to estimate edge weights since the search space of target $\mathbf{G}$ is also larger than a vector $\boldsymbol{\beta}$.

\section{Experiment Setting}
\subsection{Dataset}
We use the dataset for the CbR task released by \citet{CMRACL2019}. The dataset contains crawled articles and discussions about these articles from Reddit. The articles act as the documents, while the discussions serve as conversational contexts and responses. In total, we have 2.3M/13k/1.5k samples for training/testing/validation. % The average number of words in the document sentence and utterance on all samples are 13.78 and 18.71 respectively%; each document consists of 16.80 sentences on average.

\subsection{Implementation Details}
For all methods, we set word embedding dimension to 300 with the pre-trained GloVe~\cite{pennington2014glove}. Following \citet{CMRACL2019}, our vocabulary contains top 30k frequent tokens. We use bi-LSTMs with the hidden dimensions of 512 and the dropout rate of 0.4 in our encoders. We optimize models by Adam with an initial learning rate of 0.0005 and the batch size of 32. All conversation contexts/responses/documents are truncated to have the maximum length of 30/30/500.
%, and the document length is limited up to 500. 
For training, we set $\lambda$ as 1 in the loss of student models after tuning. 
For inference, we apply a top-$k$ random sampling decoding~\cite{edunov2018understanding} with $k$=20. The validation set is for early stopping. Aforementioned implementation details can be found in our codes \footnote{https://github.com/tianzhiliang/RAM4CbR}.

\subsection{Competing Methods}
\begin{enumerate}[wide=0\parindent,,noitemsep]
    \item \textbf{Seq2Seq}~\cite{seq2seq}. The standard Seq2Seq model that leverages only the conversational context for response generation.
    \item \textbf{MemNet}~\cite{ghazvininejad2018knowledge}. A knowledge-grounded conversation model that uses a memory network to store knowledge facts.
    \item \textbf{GLKS}~\cite{ren2019thinking}. It applies a global knowledge selector in encoding and a local selector on every decoding step. 
    \item Conversation with Machine Reading (\textbf{CMR}) \cite{CMRACL2019}. The state-of-the-art model on the CbR task, which is also our base model (Sec~\ref{sec:base_model}). Here, we use the full model of CMR (called CMR+w in \cite{CMRACL2019}), since the full model outperforms other CMR's variants on most metrics.
    We further apply the copy mechanism~\cite{see2017get} to this base model (\textbf{CMR+Copy}).
    \item Four variants of our proposed 
    models: \textbf{RAM\_T} denotes our \textbf{R}esponse-\textbf{A}nticipated \textbf{M}emory-based model with R\textbf{T}I, and \textbf{RAM\_T+Copy} denotes its copy version. \textbf{RAM\_P} and \textbf{RAM\_P+Copy} denote our model with R\textbf{P}I and its copy variant .
\end{enumerate}

\begin{table*}[htb]
%\footnotesize
%\centering
%\begin{tabular}{c|c|c|c|c|c|c|c|c|c|c}
%\hline
%\multirow{2}{*}{}     & \multicolumn{3}{c|}{Appropriateness}               & \multicolumn{3}{c|}{Grounding}                      & \multicolumn{3}{c|}{Informativeness}             &      \\ \cline{2-10}
%                      & NIST           & BLEU            & METEOR          & P       & R          & F1              & Ent-4           & Dist-1          & Dist-2          & Len  \\ \hline \hline
\centering
\scriptsize
\begin{tabular}{c|c|c|c|c|c|c|c|c|c|c|c|c|c}
\hline
\multirow{2}{*}{} & \multicolumn{3}{c|}{Appropriateness}               & \multicolumn{6}{c|}{Grounding}                      & \multicolumn{3}{c|}{Informativeness}             &      \\ \cline{2-13}
                  & NIST           & BLEU            & METEOR          & P       & R          & F1 & P$_{GT}$       & R$_{GT}$          & F1$_{GT}$             & Ent-4           & Dist-1          & Dist-2          & Len  \\ \hline \hline
RAM\_T       & 2.510 & 1.43\% & 7.74\% & \underline{4.46\%}          & \underline{0.26\%}          & \underline{0.49}\%    & \underline{1.04\%}          & \underline{0.08\%}          & \underline{0.15\%}      & \underline{9.900} & \underline{0.053} & \underline{0.290} & 15.1 \\ \hline
RAM\_P           & 2.353          & 1.40\%          & 7.59\%          & 3.89\%          & 0.21\%          & 0.41\%      & 0.97\%          & 0.07\%          & 0.13\%     & 9.891          & 0.049          & 0.279          & 14.9 \\ \hline\hline
RAM\_T (Teacher) & 2.539          & 1.43\%          & 7.85\%          & 4.47\%          & 0.26\%          & 0.49\%      & \textbf{1.05}\%          & 0.08\%          & 0.15\%    & \textbf{9.904} & 0.053          & \textbf{0.290} & 15.1 \\ \hline
RAM\_P (Teacher) & \textbf{2.551} & \textbf{1.47\%} & \textbf{7.88\%} & \textbf{4.56\%} & \textbf{0.27\%} & \textbf{0.50\%} & 0.99\%          & \textbf{0.08}\%      & \textbf{0.16\%} & 9.900          & \textbf{0.053} & 0.287          & 15.1 \\ \hline \hline
RAM\_T\_Binary           & \underline{2.560} & \underline{1.63\%} & \underline{7.91\%} & 3.75\%          & 0.21\%          & 0.40\%   & 0.87\%          & 0.07\%          & 0.12\%      & 9.890          & 0.052          & 0.283          & 15.1 \\ \hline
RAM\_P\_Binary           & 2.403  & 1.51\% & 7.63\% & 3.55\% & 0.18\% &  0.38\%  & 0.85\%          & 0.07\%          & 0.12\% & 9.887  &     0.046  &  0.274   & 14.6 \\ \hline
\end{tabular}
\caption{Performance comparison on our model variants. Line1\&2: our models trained by the full teacher-student framework. Line3\&4: our models trained with the teacher model only.
Line5\&6: our models with binary weight matrices.
%and the model with the binary weight matrix. All models are evaluated on student models by default; ``(teacher)" indicates being evaluated on teacher models. 
Bold values are the best results among the first four lines; underlines mark the best ones among the first two and last two lines. }
%}
\label{tb:model_variants}
\end{table*}

\subsection{Evaluation Metrics}
Following all metrics in \citet{CMRACL2019}, we evaluate all methods by both automatic and human evaluations. For automatic evaluations, we evaluate the responses in three aspects:
\begin{enumerate}[wide=0\parindent,noitemsep]
    \item \textbf{Appropriateness}. 
    %We evaluate the overall quality by measuring the n-gram matching between human created responses (ground-truth) and our generated ones. 
    
    We use three metrics to evaluate the overall quality of a response: \textrm{BLEU-4} \cite{papineni2002bleu}, \textrm{METEOR} \cite{banerjee2005meteor}, and \textrm{NIST} \cite{doddington2002automatic}. \textrm{NIST} is a variant of BLEU that measures n-gram precision weighted by the informativeness of n-grams.
    \item \textbf{Grounding}.  We measure the relevance between documents and generated responses
    %via token matching, 
    to reveal the effectiveness of responses exploiting the document information.
   We define $\#\mbox{overlap}$ as the number of non-stopword tokens in both the document $D$ and the generated response $\hat{R}$ but not in contexts $X$. We calculate the precision $\mbox{P}$ and recall $\mbox{R}$ as
    \begin{flalign}
    & \qquad \#\mbox{overlap}=|(D \cap \hat{R}) \backslash X \backslash S|, \\
    & \qquad \mbox{P} = \frac{\#\mbox{overlap}}{|\hat{R} \backslash S|}, \mbox{R} = \frac{\#\mbox{overlap}}{|D \backslash S|},
    \end{flalign}
    where $S$ denotes the stopword list. \textit{F1} is the harmonic mean of precision $\mbox{P}$ and recall $\mbox{R}$.

    We further propose to measure the effectiveness of exploiting the document information considering the ground-truth. In this way, we evaluate how many  ground-truth information models can exploit from the document. We define $\#\mbox{overlap}_{GT}$ as the number of non-stopword tokens in the document $D$, the generated response $\hat{R}$ and the ground-truth $R$ but not in contexts $X$. The precision and recall are as following,
    \begin{flalign}
    & \qquad \#\mbox{overlap}_{GT}=|(D \cap \hat{R} \cap R) \backslash X \backslash S|, \\
    & \qquad \mbox{P}_{GT} = \frac{\#\mbox{overlap}_{GT}}{|\hat{R} \backslash S|}, \mbox{R}_{GT} = \frac{\#\mbox{overlap}_{GT}}{|D \backslash S|},
    \end{flalign}
    
    where \textit{$F1_{GT}$} is the harmonic mean of precision \textit{$P_{GT}$} and recall \textit{$R_{GT}$}.
    
    \item \textbf{Informativeness}. \textit{Ent-n} \cite{seq2BF} measures responses' informativeness with the entropy of the n-gram count distribution. \textit{Dist-n} \cite{li2016diversitymmi} evaluates the diversity of responses via the proportion of unique n-grams among all responses.
\end{enumerate}

\noindent
For human evaluations, we hire five annotators from a commercial annotation company to evaluate 200 randomly selected test samples, and results from different models are shuffled. The annotators evaluate on a 5-point scale in three aspects: overall quality (\textit{H-Appr}), relevance with documents (\textit{H-Ground}), and informativeness (\textit{H-Info}).
%1 point for a response irrelevant to the context and document, 3 points for a valid but meaningless response, 5 points for a coherent and appropriated response without typos.
%The relevance with documents (\textit{H-Ground}): 1 point for a response irrelevant to document, 3 points for the response that is a little similar to the document, 5 points for the the response matching the document totally and sharing some entity words with document.
%The informativeness (\textit{H-Info}): 1 point for the universal response or the response containing no more than three unique tokens, 3 points for a normal response of a single clause or a single topic, and 5 points for an informative response including at least two clauses of different topics.
% , which transfer the current conversation to another scenario (For example, the query is ``How's the weather?"; and response ``It's fine today, let's play basketball" transfers the weather topic to sports, which should be marked as 3 points).
%Points of 2 and 4 are for decision dilemmas of above three metrics.

\begin{table}[t]
\centering
\footnotesize
\begin{tabular}{c|c|c|c}
\hline
            & H-Appr & H-Ground  & H-Info \\ \hline \hline
Human        &   2.986     &    2.521          &      3.007    \\ \hline
Seq2Seq     & 1.902        &    1.564     &      2.040    \\ \hline
MemNet      &    1.872     &    1.574      &      2.105    \\ \hline
GLKS      &      2.073    &     1.593   &   2.071       \\ \hline
CMR         & 2.188     & 1.678      &      2.219     \\ \hline
CMR+Copy    &   2.063       &   1.773    &      2.075    \\ \hline \hline
RAM\_T      &   \textbf{2.259} &    1.714       &      \textbf{2.312}    \\ \hline
RAM\_P       &    2.213 &    1.682      &    2.231      \\ \hline
RAM\_T+Copy    & 2.109  &    \textbf{1.861}     &    2.240      \\ \hline
RAM\_P+Copy  &     2.114  &    1.775   &     2.115     \\ \hline
\end{tabular}
\caption{Human annotation results.}
\label{tb:overall_human}
\end{table}

\begin{table}[t]
\centering
\footnotesize
%\resizebox{7.2cm}{!}{
\begin{tabular}{c|c|c|c|c}
\hline
            & \multicolumn{2}{c|}{Top10 tokens} & \multicolumn{2}{c}{Top20 tokens} \\ \cline{2-5} 
            & Emb-M        & Emb-B       & Emb-M        & Emb-B       \\ \hline
CMR         & 0.482           & 0.356           & 0.571           & 0.420           \\ \hline
RAM\_T\_Soft & \textbf{0.745}  & \textbf{0.520}  & \textbf{0.867}  & \textbf{0.616}  \\ \hline
RAM\_P\_Soft & 0.518           & 0.441           & 0.634           & 0.493           \\ \hline
\end{tabular}
%}
\caption{Similarity between important document tokens picked by gold responses and the accumulated attention weights in the models.} 
%The token number $K$ is 10 or 20.}
\label{tb:attn_effect}
\end{table}

% \begin{figure}[htb]
% 	\centering
% 	\includegraphics[width=.485\textwidth]{case_attn.png}
% 	\caption{Attention weight distribution for each tokens on TN\_S and CMR. The case comes from case1 in Table~\ref{tb:showcase}. The emphasized tokens (``num" and ``premier") in TN\_S model contribute to its responses. Due to the document length, we only show high weight tokens in both methods here.}
% 	\label{fig:case_attn}
% \end{figure}

\section{Experimental Results and Analysis}
In this part, we first show the performance of all methods in~Sec~\ref{sec:exp_overall}. Then, we validate the effectiveness of response anticipation on CbR in~Sec~\ref{sec:exp_ra} by comparing the top similar tokens with the response using their representations in the memory.  
We also compare more variants  of our model in~Sec~\ref{sec:exp_ana_variants}, including the token importance versus pairwise importance, and each method with continuous weights versus their variants with binary weights. At last, we conduct a case study
%and show how attention contributes to generated responses 
in~Sec~\ref{sec:exp_case}.

%description of the proposed teacher-student architecture for CbR, then briefly state the base model. The detailed teacher model and student model are presented in Sec.~\ref{sec:teacher_model} and ~\ref{sec:student_model}.
%Lastly, we summarize the training updates of the two models in Sec.~\ref{sec:model_training}.

\subsection{Overall Performance}
\label{sec:exp_overall}
Results of all models on automatic and human evaluations are shown in Table~\ref{tb:overall_auto} and Table~\ref{tb:overall_human}. MemNet outperforms Seq2Seq on most metrics, which validates that it is important to utilize document information in CbR. However, MemNet only slightly improves on Grounding.
% MemNet improves the Grounding slightly since it applies memory network directly on CbR can only slightly improve the .
%but applying memory network directly on CbR results in only a slightly improvement on Grounding. 
Both GLKS and CMR outperform MemNet on most metrics, indicating that it matters how to construct the document memory used in conversation models for CbR. %CMR is the state-of-the-art baseline on CbR, and 
Compared with CMR, CMR+Copy is more competitive on Grounding but weaker on other metrics.

Our proposed models outperform other competing methods on all metrics, including automatic and human evaluations. For models without the copy mechanism, RAM\_T  performs the best. %including the state-of-the-art baseline (CMR) on all metrics, 
%and RAM\_P exceeds CMR on most metrics except \textif{Dist}. 
For models with copy, RAM\_T+Copy and RAM\_P+Copy excel CMR+Copy on most metrics. Overall, our proposed strategy works well on both the model with and without copy mechanism.
We will compare RAM\_T and RAM\_P in details in Sec~\ref{sec:exp_ana_variants}. 
%As for different variants of our model, those with RTI stratey (RAM\_T, RAM\_T+Copy) works better than those with RPI (RAN\_P, RAM\_P+Copy) and we will analyze that in Sec.~\ref{sec:exp_ana_variants}. %Comparison on models with copy and its variants without copy suggests that copying from document tokens will greatly enhance the relevance to document but decrease the Appropriateness and informativeness.
% provide a systematical analysis about those strategies in Sec.~\ref{}.
% a systematical analysis in Sec.~\ref{} provides an explanation on those strategies.
%Our proposed TS\_E\_Soft is superior to CMR on most metrics except \textit{Dist} score.

\begin{figure*}[t]
	\centering
	\includegraphics[width=1\textwidth]{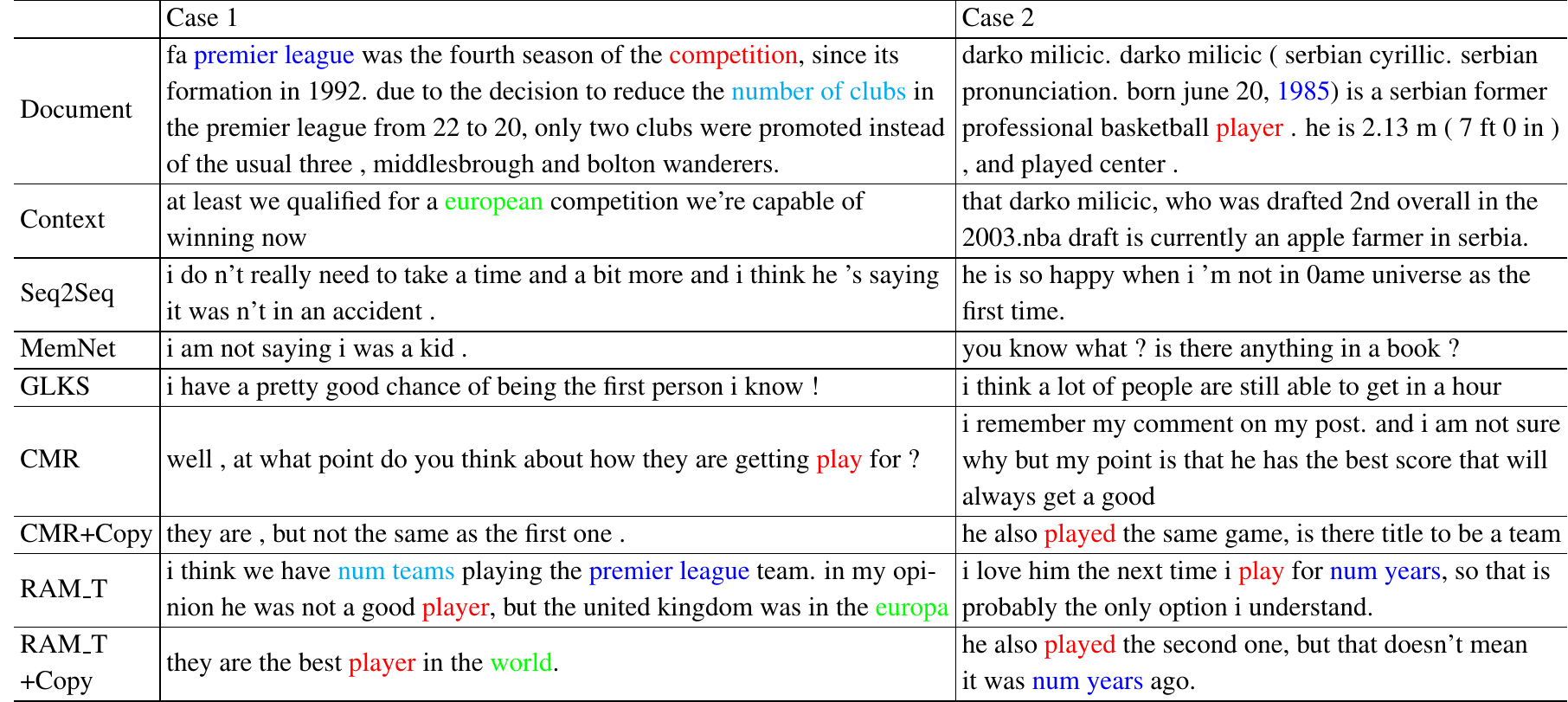}
	\caption{Test samples with generated responses of all models. A colored word in the responses indicate that it has similar words with documents or contexts, which are marked in the same color.}
	\label{fig:casestudy}
\end{figure*}

\begin{figure}[t]
	\centering
	\includegraphics[width=0.95\columnwidth]{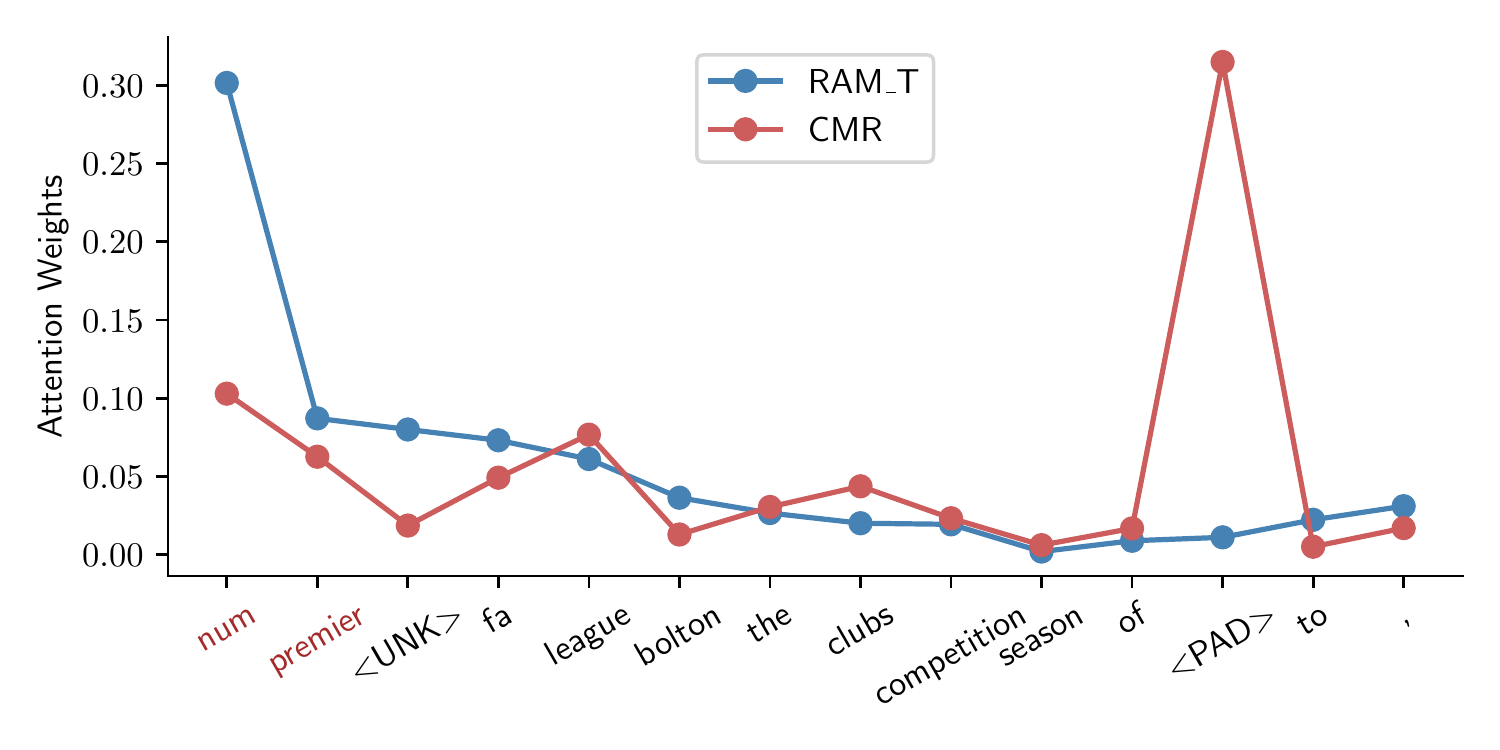}
	\caption{The accumulated attention weights of documents tokens on RAM\_T and CMR on Case 1 in Fig.~\ref{fig:casestudy}.  
	%The emphasized tokens (``num”, ``premier”) in RAM\_T contribute to its responses. 
	We only show top tokens in both methods here.}
	\label{fig:case_attn}
\end{figure}

\subsection{Effectiveness of Response Anticipation}
\label{sec:exp_ra}
% We design a metric $\mathcal{C}$ to verify whether anticipating response contributes in building document memory. The metric $\mathcal{C}$ measures the coherence between important tokens from document with respect to ground-truth and those with respect to attention distribution. We require ground-truth response to recommend a set of important document tokens $D_I$ according the semantic similarity between each document tokens and response; we also require attention weight to recommend a important token set $D'_I$ according to the summarization of attention weight on each tokens. We treat the semantic similarity between above two sets as $\mathcal{C}$.
In this section, we investigate whether anticipating response contributes to building a better document memory. 
% The metric $\mathcal{C}$ measures the coherence between important tokens from document with respect to ground-truth and those with respect to attention weights. 
%The coherence score is a measure of how well a model picks up helpful tokens and writes in memory.
%Specifically, we measure the semantic similarity between important document tokens picked by our attention and important document tokens picked by the ground-truth response. 
We first calculate the semantic similarity between each document token and the response using their Glove embeddings, and select top $K$ document tokens. 
%We also select another set of $K$ tokens according to the self-attention weights $\mathbf{A}$ in Eq.~\ref{eq:base_attn}. 
% with the top $K$ weights, %We treat the semantic similarity between the above two sets as $\mathcal{C}$.
Next, we accumulate the attention weights of each token in all attention distributions in the self-attention weights $\mathbf{A}$ in Eq.~\ref{eq:base_attn}, i.e. summation over each column of $\mathbf{A}$. 
Then we select the top $K$ tokens according to their accumulated attention weights.
Here, we set $K=10, 20$.
%To measure the two sets of tokens extracted above,
%
We apply metrics in ~\citet{liu2016not}
to calculate the similarity of two token sets extracted above, including
%measure the semantic similarity mentioned above 
%of the two sets of tokens 
maximal tokens-tokens embedding similarity (Emb-M) and bag-of-word embedding similarity (Emb-B). 
A higher similarity score indicates more response information anticipated by the model. %The ideal model, which captures response information completely, can pick a same set as the ground-truth resulting in the maximum of $\mathcal{C}$.
    Table~\ref{tb:attn_effect} shows the results of our two models RAM\_T and RAM\_P as well as CMR (We use the original self-attention matrix $\mathbf{A}$ for the above calculation for CMR). Results demonstrate that our model is able to output more response-anticipated self-attention distributions, which benefits generating a response close to the ground truth.
%picks more document tokens relevant to the response than CMR, implying our attention captures the response information well. 

%graph captures document tokens similar to responses, we conduct a comparison between original attention of CMR and attention refined by our selecting graph. To achieve it, we measure the overlap between top XX high attention weight tokens and top XX document tokens similar to the responses. 
\subsection{Analysis on Different Model Variants}
\label{sec:exp_ana_variants}
%To further verify the effectiveness of our strategies, we compare the following variants of our model.
\paragraph{Token importance vs Pairwise importance.} We compare our model variants with different strategies to construct the response-aware/anticipated weight matrix , i.e. RAM\_T (Eq.~\ref{eq:soft_node_weight}) and RAM\_P (Eq.~\ref{eq:edge_weight}).
We not only compare their overall performance by the teacher-student framework (Eq.~\ref{eq:teacher_loss} \& ~\ref{eq:student_loss}) but also the teacher model only (Eq.~\ref{eq:student_loss}). 
%in two aspects during generation (testing): generating from student models (normal usage) and generating from teacher model (access to ground-truth). 

The first four rows in Table~\ref{tb:model_variants} shows the results.
We have an interesting finding that RAM\_P underperforms RAM\_T in the full teacher-student framework, but outperforms RAM\_T on the mode with teacher model only on most metrics. This result is actually consistent with our discussion in Sec~\ref{sec:model_training} that RAM\_P has a higher capacity to carry more information in $\mathbf{G}$, thus its teacher model yields better performance. However, %in terms of joint optimization with the student model, 
for the student model, RAM\_P is more difficult to converge to a good local optimum due to more parameters to be estimated, resulting in that its overall performance may not exceed that of RAM\_T.

\paragraph{Continuous weight vs Binary weight.} 
We also compare the model variants with continuous weight (Eq.~\ref{eq:soft_node_weight}) and binary weight (Eq.~\ref{eq:hard_node_weight}). The last two rows in Table~\ref{tb:model_variants} give the results of the variants of RAM\_T and RAM\_P with a binary $\mathbf{G}$. We can see that both RAM\_T and RAM\_P with a binary weight matrix performs better on Appropriateness, which means a sparse $\mathbf{G}$ on the attention matrix can help select more concise information to construct the memory. Nevertheless, models with a continuous weight matrix can generate more informative responses owing to their ability to access broader and more information from the document.

\subsection{Case Study}
\label{sec:exp_case}
Table~\ref{fig:casestudy} shows two test samples with generated responses of all models. For Case 1, Seq2Seq and MemNet cannot generate responses relevant to either the document or context. CMR catches the topic ``sports", while GLKS and CMR+Copy use ``first person" and ``first one" to reflect ``only two" mentioned in the document. The response of RAM\_T contains information related to both document (``num teams" and ``premier league") and context (``europa"). RAM\_T+Copy is also highly relevant to the document and the context, and copies ``player" from the document. For Case 2, the first four methods have little relation to the document or the context. CMR+Copy mentions ``played". Our models mention ``played" and ``num years". By examining the cases, our method shows promising improvements over existing methods. However, generation on the CbR task is very challenging and there is still a huge space to improve.

%Table~\ref{tb:showcase} shows two concrete cases of models. For the first case, Seq2Seq and MemNet mismatch the topic of the document and contexts. CMR catches the topic "sports", while CMR+Copy uses ``first one" to respond ``only two" mentioned in document. Our TS\_N contains information related to both document (``num teams" and ``premier league") and context (``europa"). TS\_N+Copy also satisfies the document and context, and copy ``player" from document. For case2, the first three methods have little relation to the document and contexts. CMR+Copy mentions ``played" by Copying. Our models mention both ``played" and ``num years".

%To evaluate our model qualitatively, we give two cases in Table~\ref{tb:showcase}. In first case, Seq2Seq and MemNet mismatch the topic of the document and contexts. CMR catches the topic "sports", while CMR+Copy uses ``first one" to respond ``only two" mentioned in document. Response from our proposed TS\_N contains information related to both document (``num teams" and ``premier league") and context (``europa"). TS\_N+Copy also has high relevance with document and context. We can see that it copies ``player" from document to response. For case2, the first three methods have little relation to the document and contexts. CMR+Copy mentions ``played" by Copying. Our models mention both ``played" and ``num years".

We plot the accumulated attention weights of RAM\_T and CMR as in Sec~\ref{sec:exp_ra} of the document tokens
%Fig.~\ref{fig:case_attn} illustrates how attention emphasizes document tokens 
on Case 1. 
%Compared with CMR's attention, 
%As can be seen in Fig.~\ref{fig:case_attn}, 
Fig.~\ref{fig:case_attn} shows that RAM\_T's attention highlights ``num" and ``premier", and thus it generates the above words in its response.

\section{Conclusion}
Focusing on the CbR task, we propose a novel response-anticipated document memory to exploit and memorize the document information that is important in response generation. We construct the response-anticipated memory by a teacher-student framework. The teacher accesses the response and learns a response-aware weight matrix; the student learns to estimate the weight matrix in the teacher model and construct the response-anticipated document memory. We verify our model on both automatic and human evaluations and experimental results show our model obtains the state-of-the-art performance on the CbR task. 

%learning a weight matrix. To learn the matrix, we construct a teacher-student framework to detect and capture the importance of document tokens or token relations with respect to responses. 

\section{Acknowledgments}
Research on this paper was supported by Hong Kong Research Grants Council under grants 16202118 and 16212516 and Tencent AI Lab Rhino-Bird Focused Research Program (No. GF202035).

\bibliography{acl2020}
\bibliographystyle{acl_natbib}

%\appendix

%\section{Appendices}
%\label{sec:appendix}
%Appendices

\end{document}